\documentclass[twoside,journal]{IEEEtran}
\IEEEoverridecommandlockouts

\usepackage[T1]{fontenc}
\usepackage{lmodern}
\usepackage{times}

\usepackage{amsmath}
\usepackage{amssymb}
\usepackage{amsfonts}
\usepackage{amsthm}
\usepackage{mathtools}
\usepackage{bm}
\usepackage{dsfont}
\usepackage{latexsym}
\usepackage{nicefrac}
\usepackage{cases,subeqnarray}
\usepackage{extarrows}

\usepackage{graphicx}
\usepackage{epsfig}
\usepackage{epstopdf}
\usepackage{float}
\usepackage[caption=false,font=footnotesize]{subfig}
\usepackage{stfloats}
\usepackage{multirow}
\usepackage{bigstrut}
\usepackage{makecell}
\usepackage{array}
\usepackage{booktabs}
\usepackage{enumitem}

\usepackage[table]{xcolor}
\usepackage{colortbl}
\usepackage{textcomp}
\usepackage{pifont}
\usepackage{microtype}
\usepackage{setspace}

\usepackage[noadjust]{cite}
\usepackage{url}
\usepackage{xr-hyper}
\usepackage[pagebackref=false,breaklinks=true,colorlinks,citecolor=blue,linkcolor=blue,bookmarks=true]{hyperref}

\usepackage{multicol}
\usepackage{balance}
\usepackage{flushend}
\usepackage[nameinlink]{cleveref}  
\usepackage{etoc}
\usepackage{algorithm}
\usepackage{algorithmicx}
\usepackage{algpseudocode}
\usepackage{titletoc}
\usepackage{pdfpages}
\usepackage{subcaption}
\usepackage[nointegrals]{wasysym}
\usepackage[font=footnotesize]{caption}

\usepackage{tikz}
\definecolor{lime}{HTML}{A6CE39}
\DeclareRobustCommand{\orcidicon}{%
\begin{tikzpicture}
\draw[lime, fill=lime] (0,0) 
circle [radius=0.16] 
node[white] {{\fontfamily{qag}\selectfont \tiny ID}};\draw[white, fill=white] (-0.0625,0.095) 
circle [radius=0.007];\end{tikzpicture}
\hspace{-2mm}}

\foreach \x in {A, ..., Z}{%
\expandafter\xdef\csname orcid\x\endcsname{\noexpand\href{https://orcid.org/\csname orcidauthor\x\endcsname}{\noexpand\orcidicon}}
}

\hypersetup{
 colorlinks=true,
 linkcolor=black,
 filecolor=black,
 urlcolor=black,
 citecolor=blue,
}

\theoremstyle{plain}
\newtheorem{assumption}{Assumption}
\newtheorem{theorem}{Theorem}
\newtheorem{thm}{Theorem}[section]
\newtheorem{lemma}{Lemma}
\newtheorem{lemm}{Lemma}
\newtheorem{proposition}{Proposition}
\newtheorem{prop}{Proposition}
\newtheorem{corollary}{Corollary}
\newtheorem{corr}{Corollary}
\newtheorem{definition}{Definition}

\theoremstyle{remark}
\newtheorem{remark}{Remark}
\newtheorem{rem}{Remark}

\crefname{section}{Sec.}{Secs.}
\crefname{subsection}{Sec.}{Secs.}
\crefname{subsubsection}{Sec.}{Secs.}
\newcommand{\cmark}{\ding{51}}
\newcommand{\xmark}{\ding{55}}
\crefname{figure}{Fig.}{Figs.}
\crefname{table}{Table}{Tables}
\crefname{equation}{Eq.}{Eqs.}
\crefname{algorithm}{Alg.}{Algs.}
\crefname{assumption}{Assumption}{Assumptions}
\crefname{theorem}{Theorem}{Theorems}
\crefname{lemma}{Lemma}{Lemmas}
\crefname{proposition}{Proposition}{Propositions}
\crefname{corollary}{Corollary}{Corollaries}
\crefname{definition}{Definition}{Definitions}
\crefformat{equation}{#2Eq.~#1#3}
\crefmultiformat{equation}{#2Eqs.~#1#3}{ and~#2#1#3}{, #2#1#3}{ and~#2#1#3}
\Crefformat{equation}{#2Eq.~#1#3}
\Crefmultiformat{equation}{#2Eqs.~#1#3}{ and~#2#1#3}{, #2#1#3}{ and~#2#1#3}

\definecolor{lightkeycolor}{RGB}{255,240,240}
\definecolor{lightgray}{RGB}{234,234,234}

\PassOptionsToPackage{table}{xcolor}
\def\RevisionPreambleLight{1}
\ifdefined\pdfminorversion\pdfminorversion=4\fi
\ifdefined\pdfobjcompresslevel\pdfobjcompresslevel=0\fi

\makeatletter
\PassOptionsToPackage{table}{xcolor}
\@ifpackageloaded{xcolor}{}{\usepackage{xcolor}}
\@ifpackageloaded{soul}{}{\usepackage{soul}}
\@ifpackageloaded{pifont}{}{\usepackage{pifont}}
\@ifpackageloaded{tikz}{}{\usepackage{tikz}}
\makeatother
\definecolor{revisionHighlightColor}{RGB}{0,0,255}
\providecommand{\blue}[1]{}
\renewcommand{\blue}[1]{\textcolor{revisionHighlightColor}{#1}}
\providecommand{\RegisterRevisionHighlightCommands}{%
  \ifdefined\cite\soulregister\cite7\fi
  \ifdefined\ref\soulregister\ref7\fi
  \ifdefined\cref\soulregister\cref7\fi
  \ifdefined\Cref\soulregister\Cref7\fi
  \ifdefined\eqref\soulregister\eqref7\fi
  \ifdefined\textbf\soulregister\textbf7\fi
  \ifdefined\textit\soulregister\textit7\fi
  \ifdefined\emph\soulregister\emph7\fi
}
\RegisterRevisionHighlightCommands
\makeatletter
\@ifundefined{hlmaskenv}{%
  \ifdefined\RevisionDisableVisualHighlight
    \newenvironment{hlmaskenv}{}{}%
  \else
    \newsavebox{\hlmasksavebox}%
  \fi
}{}
\makeatother
\ifdefined\RevisionPreambleLight
  \def\RevisionPreambleMaybeEnd{ }
\else
  \let\RevisionPreambleMaybeEnd\relax
\fi
\RevisionPreambleMaybeEnd

\usepackage{mathtools,bm,amsmath,amssymb,amsfonts,amsthm,nicefrac,latexsym,dsfont,extarrows,bbm}
\usepackage{cases,subeqnarray}
\usepackage{algorithm,algorithmicx,algpseudocode}
\usepackage{booktabs,multirow,subcaption,graphicx,epsfig,epstopdf,float,stfloats,bigstrut,makecell,array}
\usepackage[most]{tcolorbox}
\usepackage{xcolor,color,colortbl,textcomp,pifont,microtype}
\usepackage{multicol,balance,flushend,setspace,titletoc,etoc}
\usepackage[noadjust]{cite}
\usepackage{url,xr-hyper}
\makeatletter
\@ifpackageloaded{wasysym}{}{\usepackage[nointegrals]{wasysym}}
\makeatother

\usepackage[pagebackref=false,breaklinks=true,colorlinks,bookmarks=true,linkcolor=black,citecolor=blue,urlcolor=black,filecolor=blue]{hyperref}
\makeatletter
\@ifundefined{Hy@setpdfversionfalse}{}{\Hy@setpdfversionfalse}
\makeatother
\usepackage[nameinlink]{cleveref}
\RegisterRevisionHighlightCommands

\usepackage{tikz}
\definecolor{lime}{HTML}{A6CE39}
\DeclareRobustCommand{\orcidicon}{\begin{tikzpicture}\draw[lime, fill=lime] (0,0) circle [radius=0.16] node[white] {{\fontfamily{qag}\selectfont \tiny ID}};\draw[white, fill=white] (-0.0625,0.095) circle [radius=0.007];\end{tikzpicture}\hspace{-2mm}}
\foreach \x in {A, ..., Z}{\expandafter\xdef\csname orcid\x\endcsname{\noexpand\href{https://orcid.org/\csname orcidauthor\x\endcsname}{\noexpand\orcidicon}}}

\theoremstyle{plain}

\theoremstyle{remark}

\crefname{section}{Sec.}{Secs.}
\crefname{subsection}{Sec.}{Secs.}
\crefname{subsubsection}{Sec.}{Secs.}
\crefname{figure}{Fig.}{Figs.}
\crefname{table}{Table}{Tables}
\crefname{equation}{Eq.}{Eqs.}
\crefname{algorithm}{Alg.}{Algs.}
\crefname{assumption}{Assumption}{Assumptions}
\crefname{theorem}{Theorem}{Theorems}
\crefname{lemma}{Lemma}{Lemmas}
\crefname{proposition}{Proposition}{Propositions}
\crefname{corollary}{Corollary}{Corollaries}
\crefname{definition}{Definition}{Definitions}
\crefformat{equation}{#2Eq.~#1#3}
\crefmultiformat{equation}{#2Eqs.~#1#3}{ and~#2#1#3}{, #2#1#3}{ and~#2#1#3}
\Crefformat{equation}{#2Eq.~#1#3}
\Crefmultiformat{equation}{#2Eqs.~#1#3}{ and~#2#1#3}{, #2#1#3}{ and~#2#1#3}

\definecolor{lightkeycolor}{RGB}{255,240,240}
\definecolor{lightgray}{RGB}{234,234,234}

\def\x{{\mathbf x}}

\ifdefined\pdfminorversion\pdfminorversion=4\fi
\ifdefined\pdfinclusioncopyfonts\pdfinclusioncopyfonts=1\fi
\IfFileExists{macros.tex}{

}{}

\let\RevisionPreambleLight\undefined

\def\x{{\mathbf x}}

\begin{document}

\title{Continual Learning with Elastic Regularization and Synthetic Replay for Federated MLLM Fine-Tuning}

\author{%
Jing~Liu,~%
Chenxuanyin~Zou,~%
Jiayang~Ren,~%
Gaoyun~Fang,~%
Chengfang~Li,~%
Yan~Wang,~%
Zhenchao~Ma,~%
and~Bo~Hu~%
\IEEEcompsocitemizethanks{%
\IEEEcompsocthanksitem This work is supported in part by the National Natural Science Foundation of China under Grant No. 62406075, the National Key Research and Development Program of China under Grant Nos. 2023YFC3604802 and 2024YFE0200700 (Subject No. 2024YFE0200703), the Shanghai Key Technology R\&D Program under Grant No. 25511107200, and the Mitacs Project under Grant No. IT44479. (Corresponding author: Zhenchao Ma).
\IEEEcompsocthanksitem J.~Liu is with the Department of Electrical and Computer Engineering, The University of British Columbia, Vancouver, BC V6T 1Z4, Canada, also with the College of Future Information Technology, Fudan University, Shanghai 200433, China, and also with the Division of Natural and Applied Sciences, Duke Kunshan University, Kunshan 215316, China (e-mail: jing.liu@ieee.org).
\IEEEcompsocthanksitem C.~Zou and J.~Ren are with the Department of Chemical and Biological Engineering, The University of British Columbia, Vancouver, BC V6T 1Z4, Canada (e-mails: \{zcxy,~rjy12307\}@mail.ubc.ca).
\IEEEcompsocthanksitem G.~Fang is with the Dyson School of Design Engineering, Royal College of Science, Imperial College London, South Kensington Campus, London SW7 2AZ, United Kingdom (e-mail: p.fang23@imperial.ac.uk).
\IEEEcompsocthanksitem C.~Li is with the the Suzhou Institute of Biomedical Engineering and Technology (SIBET), Chinese Academy of Sciences, Suzhou 215163, China (e-mail: licf@sibet.ac.cn).
\IEEEcompsocthanksitem Y.~Wang is with the School of Data Science and Engineering, East China Normal University, Shanghai 200062, China (e-mail: yanwang@dase.ecnu.edu.cn).
\IEEEcompsocthanksitem Z.~Ma is with the Department of Electrical and Computer Engineering, The University of British Columbia, Vancouver, BC V6T 1Z4, Canada (e-mail: zhenchaoma@ece.ubc.ca).
\IEEEcompsocthanksitem B.~Hu is with the College of Future Information Technology, Fudan University, Shanghai 200433, China (e-mail: bohu@fudan.edu.cn).
}%
}

\markboth{IEEE JOURNAL OF SELECTED TOPICS IN SIGNAL PROCESSING}%
{Liu \MakeLowercase{\textit{et al}}: Continual Learning for Federated MLLM Fine-Tuning}

\IEEEtitleabstractindextext{
\begin{abstract}
Federated fine-tuning of Multimodal Large Language Models (MLLMs) across distributed networks enables privacy-sensitive adaptation to evolving social-event streams, yet a fundamental obstacle prevents robust deployment: catastrophic forgetting, wherein sequential task updates erase previously acquired knowledge across visual, linguistic, and cross-modal representations. Addressing this challenge is especially important for decentralized content moderation and crisis-understanding systems, where reliable retention of prior event knowledge underpins stable operation. To overcome this, we propose Federated Continual Multimodal Learning (\texttt{FedCMM}), a framework that embeds continual-learning safeguards into the federated optimization loop at three complementary levels. At the parameter level, modality-aware elastic weight consolidation computes separate Fisher information matrices for the vision encoder, language backbone, and cross-modal projector, providing granular, asymmetry-aware protection against modality-specific forgetting. At the data level, each client trains a lightweight local generative replay module to synthesize raw-data-free embedding-level multimodal replay tuples without any raw data sharing. At the aggregation level, Task-similarity-aware gradient aggregation adaptively filters and reweights client updates by gradient cosine similarity, suppressing conflicting directions and stabilizing the global learning trajectory. Extensive experiments on two benchmarks demonstrate that \texttt{FedCMM} consistently outperforms recent baselines on accuracy and backward transfer, confirming that coordinated replay, consolidation, and aggregation control enable robust continual adaptation across heterogeneous networked multimodal deployments.
\end{abstract}

}

\maketitle
\IEEEdisplaynontitleabstractindextext
\IEEEpeerreviewmaketitle

\graphicspath{{./imgs/}{../}}

\section{Introduction}
\label{sec:introduction}

{T}{he} proliferation of Multimodal Large Language Models (MLLMs) has marked a significant milestone in artificial intelligence, demonstrating strong capabilities in understanding and generating content that integrates vision and language~\cite{alayrac2022flamingo,liu2026edgecloud}. Driven by these capabilities, MLLMs are increasingly relevant to privacy-sensitive social-media applications such as content moderation and crisis understanding in distributed network settings~\cite{kiela2020hateful,alam2018crisismmd,liu2025networking}. However, deploying and maintaining MLLMs in these domains necessitates training methodologies that do not rely on centralized data storage. Federated Learning (FL) has emerged as a leading paradigm for privacy-sensitive machine learning, enabling collaborative model training on decentralized data sources without exposing raw user data~\cite{mcmahan2017communication,kairouz2021advances,ribero2024fl}. Consequently, the federated fine-tuning of MLLMs is a relevant step toward adaptable multimodal systems that can be updated continuously in dynamic network environments~\cite{wu2025survey}.

Reliable continual adaptation also benefits from advances in scalable optimization and stable learning, as GPU-accelerated optimal tree learning has clearly shown that scalable computation can reliably preserve strong discriminative decision quality on large datasets~\cite{ren2023gpuaccelerated}. Interpretable soft-sensor analysis emphasizes that practical learning systems should remain transparent, stable, and deployable under operational constraints~\cite{cao2025comprehensive}. Online causal discovery under parametric uncertainty provides another useful perspective, because continual systems must track changing relationships while maintaining reliable throughput~\cite{cao2026causal}.

A fundamental challenge arises when these decentralized data streams are non-stationary, evolving over time as new concepts, tasks, or data distributions emerge. In such scenarios, MLLMs must be updated sequentially to stay current, a process known as continual learning that is fraught with the peril of catastrophic forgetting, wherein the model's performance on previously learned tasks deteriorates drastically upon learning new ones~\cite{parisi2019continual,wang2024comprehensive}. A comprehensive taxonomy of continual learning strategies has been established~\cite{dong2024federated}, yet most approaches were designed for centralized settings and do not address the federated, multimodal regime. The problem is particularly acute in social-media safety and crisis-response settings: a model proficient at detecting known forms of hateful content~\cite{kiela2020hateful} must be updated to recognize new evolving patterns, while a disaster-response classifier must retain prior event categories as new crisis types emerge~\cite{alam2018crisismmd}. Addressing catastrophic forgetting within a federated framework is therefore a practical requirement for dependable continual multimodal adaptation rather than a purely conceptual exercise.

The intersection of continual learning and federated learning introduces a complex set of technical barriers. At its core, catastrophic forgetting stems from the stability-plasticity dilemma, where a model must be plastic enough to acquire new knowledge yet stable enough to retain old information~\cite{mermillod2013stability}. Foundational continual learning methods developed for centralized settings, such as regularization-based approaches like Elastic Weight Consolidation (EWC)~\cite{kirkpatrick2017overcoming} and Synaptic Intelligence~\cite{zenke2017continual}, or replay-based strategies that rehearse on stored exemplars~\cite{lopez2017gradient}, offer partial solutions. However, their direct application in FL is problematic: regularization methods often struggle to scale in the complex parameter spaces of MLLMs~\cite{greidi2024sparse}, while replay-based methods conflict with the core privacy principles of FL if raw data is stored or shared. Moreover, the inherent statistical heterogeneity (non-IID data) across clients in FL exacerbates forgetting, as divergent local updates can destructively interfere with one another during aggregation~\cite{zhao2018federated}. Communication-efficient collaborative learning and merit-gated collaborative learning further show that distributed intelligence must balance model quality with communication and system constraints~\cite{cho2023commeff}. Related clustering studies~\cite{ren2022global,ren2026global} by Ren et al. have clearly shown that representation organization can maintain principled global structure under massive data volumes, providing strong and valuable support for coherent aggregation across heterogeneous clients.

Practical implementation challenges further compound these issues, particularly for MLLMs. A key difficulty lies in the multi-faceted nature of knowledge within these models: information is encoded not just within the vision encoder and language model separately but also in the intricate cross-modal alignment modules, which introduce unique vulnerability to modality-specific forgetting and adversarial drift~\cite{lai2025badmfl}. Multimodal medical surveys and diffusion-guided semantic consistency studies reinforce the value of cross-modal alignment when visual and textual evidence carry complementary information~\cite{liu2025multimodala,liu2026diffusionguided}. Image restoration and stereo super-resolution studies provide related evidence that visual evidence quality and feature fusion remain important when downstream multimodal systems rely on image representations collected under imperfect conditions~\cite{ma2024enhancing,ma2026stereomamba}. Semantic prototype learning and people-centric domain generalization further indicate that human-centered multimodal models should retain stable discriminative behavior under domain shifts~\cite{liu2026semantic,liu2025domain}. A continual learning strategy must therefore be modality-aware, protecting critical parameters across all three components without stifling the model's ability to learn new cross-modal relationships. Furthermore, client drift becomes substantially more pronounced in a continual learning context~\cite{serra2024federated}: in a given round, some clients might be training on a new task while others remain on older data, producing gradient updates that are misaligned in their objectives. Adaptive server-side optimization methods~\cite{reddi2021adaptive,karimireddy2020scaffold} reduce drift in static settings but are not designed for the evolving task objectives of continual learning. Automated soft-sensor design offers a parallel view of deployment-facing model construction, where adaptation and monitoring must be coordinated rather than treated as isolated steps~\cite{cao2025novel}. Client-side generative replay~\cite{shin2017continual} can partially alleviate data-level forgetting, yet synchronizing it across tasks in a privacy-preserving federated regime remains an open challenge.

Recent advancements in Federated Continual Learning (FCL) have begun to tackle some of these issues. Methods such as FedWeIT~\cite{yoon2021fedweit} and Learning without Forgetting (LwF)~\cite{li2017learning} adapt knowledge distillation techniques to the federated setting, using the global model's outputs on new data as soft labels to preserve old knowledge. Progressive curriculum-based distillation provides a related example of staged supervision for stabilizing knowledge transfer~\cite{liu2026collaborative}. Others directly apply regularization by computing parameter importance on the server, while FedProx-style methods~\cite{li2020proximal} penalize deviations from the global model. More recent studies have expanded the landscape: AF-FCL selectively reuses prior knowledge under heterogeneous task drift~\cite{wuerkaixi2024accurate}, FedCBDR improves replay balance for federated class-incremental learning~\cite{qi2025fedcbdr}, cooperative multi-model training across heterogeneous devices addresses system heterogeneity from a signal-processing perspective~\cite{xu2024compfl}, and Sec-MMFL studies privacy leakage and modality-specific protection in multimodal federated learning~\cite{xiao2026enhancing}. Although these methods mark clear progress, they are still not designed for continual federated adaptation of large MLLMs with modality-aware consolidation, privacy-preserving multimodal replay, and aggregation-time conflict suppression handled in one unified pipeline.

The limitations of existing methods motivate a new approach that synergistically integrates solutions to the core challenges of multimodal forgetting, federated data heterogeneity, and privacy exposure. Our key insight is that an effective continual federated optimization framework for networked MLLMs must operate at three distinct levels. At the parameter level, modality-aware regularization protects established knowledge through parameter importance estimation~\cite{kirkpatrick2017overcoming} applied to parameter-efficient adapter representations~\cite{hu2021lora}. At the data level, a raw-data-free rehearsal mechanism actively combats forgetting, extending exemplar replay~\cite{rebuffi2017icarl} to a local generative synthesis paradigm suitable for federated LLM adaptation~\cite{ye2024openfedllm}. At the aggregation level, a similarity-driven adaptive strategy mitigates interference from heterogeneous client updates, moving beyond static parameter isolation~\cite{rusu2016progressive} toward gradient-informed coordination. By combining these three pillars, it becomes possible to balance stability and plasticity in distributed, evolving multimodal task streams such as federated social-event understanding.

To address these challenges, we propose Federated Continual Multimodal Learning (\texttt{FedCMM}), a federated continual optimization framework designed to help MLLMs learn a sequence of tasks while mitigating catastrophic forgetting. The main contribution lies in integrating three complementary mechanisms within one multimodal federated continual pipeline rather than claiming that each mechanism is independently new in isolation. MA-EWC protects modality-specific adapter parameters for the vision encoder, language backbone, and cross-modal projector, PPFR provides raw-data-free embedding-level rehearsal on each client, and TSGA reduces inter-client interference during server aggregation. Their coordinated use is what enables stable adaptation under simultaneous task drift, multimodal coupling, and client heterogeneity.

Our work offers several key contributions:
\begin{itemize}[leftmargin=*]
    \item We propose \texttt{FedCMM}, a federated continual multimodal learning framework designed to mitigate catastrophic forgetting during privacy-sensitive sequential MLLM fine-tuning.
    \item We introduce three complementary components: MA-EWC for modality-specific parameter protection, raw-data-free federated replay for synthetic rehearsal, and TSGA for similarity-aware aggregation.
    \item We conduct extensive experiments on two challenging continual federated benchmarks that show that \texttt{FedCMM} outperforms strong federated continual learning baselines.
\end{itemize}

The remainder of this paper is organized as follows. \cref{sec:related_work} reviews related work. \Cref{sec:preliminaries,sec:method} present the technical details of our proposed \texttt{FedCMM} framework. \cref{sec:experiments} presents the experimental setup, main results, and analysis. Finally, \cref{sec:conclusion} concludes the paper and discusses future work.

\section{Related Work}
\label{sec:related_work}

Adjacent studies on multimodal, distributed, and people-centric learning further motivate the setting considered here. Surveys on diffusion-based anomaly detection highlight the importance of robust representations when abnormal events evolve over time~\cite{liu2025anomaly,liu2025survey}.

\subsection{Regularization and Replay in Continual Learning}
Continual Learning (CL) seeks to preserve prior knowledge while adapting to new data streams~\cite{parisi2019continual,mermillod2013stability}. Existing CL methods mainly rely on regularization, replay, or parameter isolation. Regularization-based approaches such as EWC and SI constrain updates to important parameters~\cite{kirkpatrick2017overcoming,zenke2017continual}, but can weaken under large task shifts. Replay-based methods such as GEM and iCaRL rehearse past knowledge effectively~\cite{lopez2017gradient,rebuffi2017icarl}, yet storing raw exemplars is difficult to justify in privacy-sensitive deployments. Parameter-isolation methods reduce interference by reserving task-specific subspaces~\cite{rusu2016progressive}, although their scalability remains limited. These foundations are useful, but they were developed for centralized training and do not by themselves resolve multimodal continual adaptation under federated communication constraints.

\subsection{Optimization Under Heterogeneity in Federated Learning}
FL trains models across decentralized data sources without centralizing raw user data~\cite{kairouz2021advances,mcmahan2017communication}. Its main optimization difficulty is statistical heterogeneity, which induces client drift and unstable aggregation~\cite{zhao2018federated}. Representative remedies include client-side regularization such as FedProx~\cite{li2020proximal} and server-side adaptive aggregation such as FedAdam, FedYogi, and SCAFFOLD~\cite{reddi2021adaptive,karimireddy2020scaffold}. These methods improve robustness on static non-IID distributions, but they assume a fixed task objective and therefore do not directly address temporal drift in continual multimodal streams.

\subsection{Federated Continual Learning Under Task Drift}
Federated Continual Learning (FCL) combines continual adaptation with decentralized optimization. Prior work mainly ports centralized CL ideas into FL, including distillation-based approaches such as LwF-FL and FedWeIT~\cite{li2017learning,yoon2021fedweit}, EWC-style consolidation~\cite{kirkpatrick2017overcoming}, and privacy-aware generative replay variants~\cite{shin2017continual,wuerkaixi2024accurate,qi2025fedcbdr}. Recent studies also examine harsher settings with online drift and asynchronous task progression~\cite{serra2024federated}. Even with these advances, current FCL evidence is dominated by unimodal benchmarks or partial solutions that target only one source of failure at a time. The remaining gap for federated MLLM adaptation is therefore integrative: multimodal continual learning requires coordinated replay, modality-aware consolidation, and aggregation-time conflict control within the same training pipeline rather than a single mechanism alone.

\section{Preliminaries}
\label{sec:preliminaries}

FL enables model training on decentralized data without requiring data centralization. Consider a federated network of $K$ clients, indexed by $k \in \{1, \dots, K\}$. Each client $k$ possesses a local dataset $\mathcal{D}_k = \{(\mathbf{x}_i, y_i)\}_{i=1}^{N_k}$, where $N_k = |\mathcal{D}_k|$ is the number of local samples. The total number of samples in the network is $N = \sum_{k=1}^K N_k$. The goal of FL is to collaboratively train a single global model with parameters $\mathbf{w}$ by minimizing a global objective function $L(\mathbf{w})$, which is a weighted average of the local client objectives $L_k(\mathbf{w})$:
\begin{align}
\min_{\mathbf{w}} L(\mathbf{w}) = \sum_{k=1}^K \frac{N_k}{N} L_k(\mathbf{w}),
\label{eq:fl_objective}
\end{align}
where $L_k(\mathbf{w}) = \frac{1}{N_k} \sum_{(\mathbf{x}_i, y_i) \in \mathcal{D}_k} \ell(f(\mathbf{x}_i; \mathbf{w}), y_i)$ is the local loss on client $k$'s data, with $f(\cdot; \mathbf{w})$ representing the model's prediction and $\ell(\cdot, \cdot)$ being a task-specific loss function. The canonical FL algorithm, FedAvg~\cite{mcmahan2017communication}, iteratively performs the following steps: (1) a central server broadcasts the current global model $\mathbf{w}$ to a subset of clients; (2) each selected client performs multiple local training steps on its data to obtain updated local parameters $\mathbf{w}_k$; (3) the clients send their updated parameters back to the server, which aggregates them to produce the next global model $\mathbf{w} \leftarrow \sum_{k} \frac{N_k}{N} \mathbf{w}_k$.

CL, also called lifelong learning, addresses the challenge of training a model on a sequence of tasks over time without access to the full data from previous tasks. Formally, the model is exposed to a sequence of $T$ tasks, $\mathcal{T} = \{\mathcal{T}_1, \mathcal{T}_2, \dots, \mathcal{T}_T\}$, where each task $\mathcal{T}_j$ has an associated data distribution $P^{(j)}(\mathbf{x}, y)$. The primary challenge in CL is catastrophic forgetting, where the model's performance on earlier tasks $\{\mathcal{T}_1, \dots, \mathcal{T}_{j-1}\}$ degrades significantly after being trained on the current task $\mathcal{T}_j$. The objective of CL is to learn the parameters $\mathbf{w}^{(T)}$ for the model after the final task $\mathcal{T}_T$ such that the model performs well across all tasks seen so far. A successful CL method must balance plasticity, the ability to learn new knowledge from the current task, with stability, the ability to retain knowledge from past tasks.
The federated continual multimodal learning setting integrates the challenges of FL and CL in the context of MLLMs. In this setting, a sequence of $T$ distinct multimodal tasks arrives over time. The data for each task $\mathcal{T}_j$ is distributed across the $K$ federated clients, where each client $k$ holds a local dataset $\mathcal{D}_k^{(j)}$. Each data sample is a multimodal pair $(\mathbf{x}_i, y_i)$, where the input $\mathbf{x}_i = (\mathbf{v}_i, \mathbf{t}_i)$ consists of a visual component $\mathbf{v}_i \in \mathcal{V}$ (e.g., an image) and a textual component $\mathbf{t}_i \in \mathcal{T}$ (e.g., a text prompt). The model $f(\cdot; \mathbf{w})$ is an MLLM parameterized by $\mathbf{w}$.

When the system is learning task $\mathcal{T}_j$, it is assumed that data from previous tasks $\{\mathcal{T}_1, \dots, \mathcal{T}_{j-1}\}$ is no longer directly available, and client data cannot be shared with the server or other clients due to privacy constraints. The global objective at the end of learning task $\mathcal{T}_T$ is to find optimal model parameters $\mathbf{w}^*$ that minimize the average loss across all tasks encountered:
\begin{align}
\mathbf{w}^* = \arg\min_{\mathbf{w}} \frac{1}{T} \sum_{j=1}^T L^{(j)}(\mathbf{w}),
\label{eq:fcml_objective}
\end{align}
where $L^{(j)}(\mathbf{w}) = \sum_{k=1}^K \frac{N_k^{(j)}}{N^{(j)}} L_k^{(j)}(\mathbf{w})$ is the global loss for task $\mathcal{T}_j$. The core challenge is to minimize the objective in \cref{eq:fcml_objective} under the dual constraints of (1) not having direct access to past task data and (2) adhering to the privacy-preserving communication protocol of federated learning.

\section{Method}
\label{sec:method}

\begin{figure*}[!t]
    \centering
    \includegraphics[width=\textwidth]{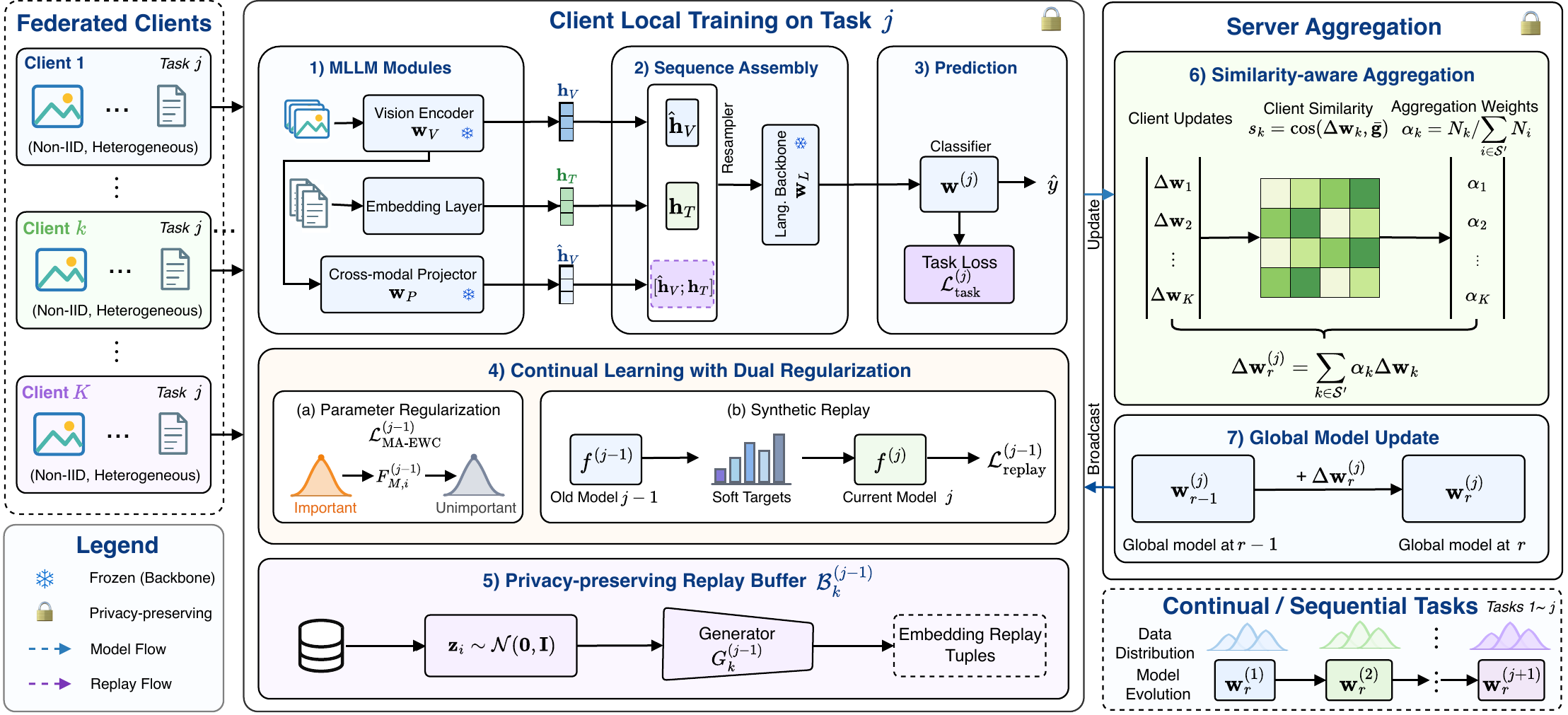}
    \caption{Overview of the \texttt{FedCMM} framework for privacy-preserving federated continual multimodal learning. For task $\mathcal{T}_j$ and communication round $r$, the server broadcasts $\mathbf{w}^{(j-1)}_{r-1}$ to selected clients. Each client trains on $\mathcal{D}_k^{(j)} \cup \mathcal{B}_k^{(j-1)}$, uses MA-EWC to protect modality-specific parameters $\mathbf{w}_V$, $\mathbf{w}_L$, and $\mathbf{w}_P$, and uploads the local update $\Delta\mathbf{w}_k$. The server then computes TSGA scores $s_k$, filters conflicting updates, aggregates accepted clients with weights $\alpha_k$ to form $\Delta\mathbf{w}^{(j)}_{r}$, and broadcasts the updated global model for the next round and task, with replay buffers and continual task evolution maintained locally.}
    \label{fig:framework_overview}
\vspace{-15px}
\end{figure*}

\subsection{Method Overview}
To address the challenge of catastrophic forgetting in federated continual learning for MLLMs, we propose \texttt{FedCMM}, a framework that integrates three synergistic components to balance knowledge stability and plasticity. As illustrated in \cref{fig:framework_overview}, \texttt{FedCMM} embeds continual-learning safeguards directly into the federated training loop while keeping the dense MiniCPM-V backbone frozen and updating only parameter-efficient LoRA adapters. The trainable parameter groups are denoted by $\mathbf{w}_V$, $\mathbf{w}_L$, and $\mathbf{w}_P$, corresponding to the vision encoder, language backbone, and cross-modal projector adapters, respectively. Each client combines current-task data with synthetic replay samples, constrains important modality-specific parameters through MA-EWC, and uploads only its local model update. The server then evaluates update consistency through TSGA, suppresses conflicting directions, and broadcasts the aggregated global model for the next round and task. Their dominant roles differ across failure modes: PPFR contributes most when old decision boundaries require rehearsal support, MA-EWC becomes more important when task switches induce module-specific adapter drift, and TSGA becomes more important when heterogeneous client updates are directionally inconsistent at aggregation time.
Formally, when learning the $j$-th task $\mathcal{T}_j$, each participating client $k$ aims to find local model parameters $\mathbf{w}_k^{(j)}$ that minimize a composite loss function on its local data $\mathcal{D}_k^{(j)}$ and a set of locally generated embedding-level replay tuples $\mathcal{B}_k^{(j-1)}$ from the previous task. The local objective for client $k$ is:
\begin{align}
\resizebox{0.9\columnwidth}{!}{$\displaystyle
\mathcal{L}_k^{(j)}(\mathbf{w}) = \mathcal{L}_{\text{task}}^{(j)}(\mathbf{w}) + \gamma\,\mathcal{L}_{\text{replay}}^{(j-1)}(\mathbf{w}) + \lambda\,\mathcal{L}_{\text{MA-EWC},k}^{(j-1)}(\mathbf{w})
$}
\label{eq:overall_local_loss}
\end{align}
where $\mathcal{L}_{\text{task}}^{(j)}$ is the supervised loss on current-task real samples, $\mathcal{L}_{\text{replay}}^{(j-1)}$ is the embedding-level rehearsal loss (see \cref{eq:replay_loss}), $\gamma$ weights the replay contribution, and $\lambda$ balances the MA-EWC regularization. In our implementation, $\gamma$ is fixed to $0.5$ on both PHEME and CrisisMMD so that replay remains a stable auxiliary signal rather than overwhelming the current-task supervision, whereas $\lambda$ controls how aggressively past-task parameter importance is preserved. After local training, the server aggregates the resulting client updates using the TSGA mechanism to produce the new global model $\mathbf{w}^{(j)}$.
For clarity, we explicitly partition the trainable parameters as
\begin{align}
\resizebox{0.9\columnwidth}{!}{$\displaystyle
\mathbf{w} = \mathbf{w}_V \cup \mathbf{w}_L \cup \mathbf{w}_P,\quad
\mathbf{w}_V \cap \mathbf{w}_L = \mathbf{w}_V \cap \mathbf{w}_P = \mathbf{w}_L \cap \mathbf{w}_P = \emptyset
$}
\label{eq:parameter_partition}
\end{align}
where the three subsets correspond to the vision encoder, the language backbone, and the cross-modal projector. In the MLLM forward pass, the vision encoder extracts visual representations $\mathbf{h}_V$, which are mapped by the cross-modal projector into the language embedding space as projected tokens $\hat{\mathbf{h}}_V$. Specifically, the projected tokens are concatenated with text token embeddings $\mathbf{h}_T$ and jointly processed by the language backbone to produce the output logits. Since we employ Low-Rank Adaptation (LoRA)~\cite{hu2021lora} for parameter-efficient fine-tuning (PEFT), $\mathbf{w}_V$, $\mathbf{w}_L$, and $\mathbf{w}_P$ refer specifically to the trainable LoRA adapter parameters within each module, while the dense pre-trained backbone weights are kept frozen. The task and replay contributions to \cref{eq:overall_local_loss} are:
\begin{align}
\mathcal{L}_{\text{task}}^{(j)}(\mathbf{w}) &= \mathbb{E}_{(\mathbf{x}, y) \sim \mathcal{D}_k^{(j)}} [\ell(f(\mathbf{x}; \mathbf{w}), y)],
\label{eq:task_loss}
\\
\mathcal{L}_{\text{replay}}^{(j-1)}(\mathbf{w}) &= \resizebox{0.6\hsize}{!}{$\displaystyle \mathbb{E}_{(\hat{\mathbf{h}}_V, \hat{\mathbf{h}}_T, \hat{y}) \sim \mathcal{B}_k^{(j-1)}} [\ell(f_{\mathrm{emb}}(\hat{\mathbf{h}}_V, \hat{\mathbf{h}}_T; \mathbf{w}), \hat{y})],$}
\label{eq:replay_loss}
\end{align}
Here $f(\mathbf{x};\mathbf{w})$ is the standard MLLM forward on raw image-text inputs, while $f_{\mathrm{emb}}(\hat{\mathbf{h}}_V, \hat{\mathbf{h}}_T;\mathbf{w})$ injects the generated embeddings directly at the visual-token and text-prompt level, bypassing the input encoder; $\gamma$ weights the replay contribution and is kept fixed across datasets to simplify reproduction.

\subsection{Modality-Aware Elastic Weight Consolidation}
A primary challenge in applying continual learning techniques to MLLMs is the complex and modular nature of their architecture. Knowledge is distributed across a vision encoder, a language model, and a cross-modal projector that aligns the two modalities. Standard regularization methods like Elastic Weight Consolidation (EWC)~\cite{kirkpatrick2017overcoming} treat all model parameters uniformly. A uniform penalty is suboptimal because it overlooks that the importance of parameters can vary significantly across different modalities. Forgetting might occur in the visual understanding part while the linguistic part remains stable, or vice-versa.

To address this, we introduce Modality-Aware EWC (MA-EWC), which applies a more granular regularization by treating each core module of the MLLM independently. The total set of trainable parameters $\mathbf{w}$ is partitioned into three disjoint subsets: vision parameters $\mathbf{w}_V$, language parameters $\mathbf{w}_L$, and cross-modal projector parameters $\mathbf{w}_P$. For each subset, client $k$ computes a separate diagonal Fisher Information Matrix (FIM) at the end of learning task $\mathcal{T}_{j-1}$. The FIM serves as a proxy for parameter importance, with larger diagonal values indicating parameters that are more critical for performance on past tasks. The client-specific FIM for a parameter set $\mathbf{w}_M$ (where $M \in \{V, L, P\}$) is approximated as:
\begin{align}
\resizebox{0.9\columnwidth}{!}{$\displaystyle
\mathbf{F}_{M,k}^{(j-1)} = \mathbb{E}_{(\mathbf{x}, y) \sim \mathcal{D}_k^{(j-1)}} \left[ \left( \nabla_{\mathbf{w}_M} \log p(y|\mathbf{x}; \mathbf{w}_M^{(j-1)*}) \right)^2 \right]
$},
\label{eq:fim}
\end{align}
At the element level, the diagonal approximation is accumulated sample-wise as
\begin{align}
\resizebox{0.9\columnwidth}{!}{$\displaystyle
F_{M,k,i}^{(j-1)} \approx \frac{1}{|\mathcal{D}_k^{(j-1)}|} \sum_{(\mathbf{x}, y) \in \mathcal{D}_k^{(j-1)}} \left(\frac{\partial \log p(y \mid \mathbf{x}; \mathbf{w}_M^{(j-1)*})}{\partial w_{M,i}}\right)^2
$},
\label{eq:fim_element}
\end{align}
where $\mathbf{w}_M^{(j-1)*}$ are the optimal parameters for module $M$ after task $\mathcal{T}_{j-1}$. In the federated setting, each client $k$ independently computes and stores its own modular FIM $\mathbf{F}_{M,k}^{(j-1)}$ over its local dataset $\mathcal{D}_k^{(j-1)}$ and retains the corresponding reference parameters $\mathbf{w}_{M,k}^{(j-1)*}$, consistent with the per-client notation used in \cref{alg:fedcmm_server}. The MA-EWC regularization loss is then the sum of the penalties for each module:
\begin{align}
\resizebox{0.85\columnwidth}{!}{$\displaystyle
\mathcal{L}_{\text{MA-EWC},k}^{(j-1)}(\mathbf{w}) = \frac{1}{2} \sum_{M \in \{V, L, P\}} \sum_{i \in \mathbf{w}_M} F_{M,k,i}^{(j-1)} (w_i - w_i^{(j-1)*})^2
$},
\label{eq:ma_ewc_loss}
\end{align}
which can be rewritten as the sum of modality-specific penalties:
\begin{align}
\resizebox{0.85\columnwidth}{!}{$\displaystyle
\mathcal{L}_{\text{MA-EWC},k}^{(j-1)}(\mathbf{w}) =
\mathcal{L}_{V,k}^{(j-1)}(\mathbf{w}_V) +
\mathcal{L}_{L,k}^{(j-1)}(\mathbf{w}_L) +
\mathcal{L}_{P,k}^{(j-1)}(\mathbf{w}_P)
$},
\label{eq:maewc_decomposition}
\end{align}
where
\begin{align}
\mathcal{L}_{M,k}^{(j-1)}(\mathbf{w}_M) = \frac{1}{2}\sum_{i \in \mathbf{w}_M} F_{M,k,i}^{(j-1)} (w_i - w_i^{(j-1)*})^2.
\label{eq:module_ewc}
\end{align}
where $F_{M,k,i}^{(j-1)}$ is the $i$-th diagonal element of the client-specific FIM for module $M$. In deployment, each client retains one modular diagonal-Fisher snapshot together with the corresponding adapter reference parameters from the most recently completed task, and refreshes that snapshot at the next task boundary rather than storing an ever-growing bank of per-task matrices. Because MA-EWC tracks only LoRA adapter parameters while the dense backbone remains frozen, its storage cost scales with the active PEFT state instead of the full MLLM parameter count. Importantly, the Fisher matrix and its diagonal snapshot are computed and stored purely locally, requiring no transmission between client and server. By separating the FIM computation, MA-EWC can, for example, heavily penalize changes to critical vision parameters if the new task is visually similar to a past one, while allowing more flexibility in the language parameters if the new task introduces novel textual concepts. A modality-specific approach thus provides a more effective safeguard against the nuanced forms of forgetting.

\subsection{Privacy-Preserving Federated Replay}
Rehearsal, or replaying exemplars from past tasks, is one of the most effective strategies for mitigating catastrophic forgetting. However, in the federated setting, storing and replaying raw data from previous tasks is infeasible due to privacy constraints and storage limitations on client devices. To overcome this, we propose Privacy-Preserving Federated Replay, a mechanism that leverages client-side generative models to create synthetic embedding-level replay tuples while avoiding raw-data exchange.
After a client $k$ completes its training on task $\mathcal{T}_j$, it uses its local dataset $\mathcal{D}_k^{(j)}$ to train a lightweight, local replay generator $G_k^{(j)}$. Importantly, PPFR does not attempt to generate raw image pixels or free-form text tokens. Instead, it operates in the frozen MLLM representation space and synthesizes multimodal replay tuples consisting of a visual-token embedding, a text-prompt prototype embedding, and a replay label. In our implementation, $G_k^{(j)}$ is a compact latent-noise module driven by Gaussian codes $\mathbf{z}\sim\mathcal{N}(\mathbf{0},\mathbf{I})$, a 64-dimensional latent space, and an event-conditioned code. A two-layer visual decoder maps the latent code to the MiniCPM visual-token embedding space, while a prompt-prototype table with a small residual MLP produces the corresponding textual embedding. The replay label is obtained from the local event label during fitting and stored as a teacher-smoothed soft target.

The generator is trained locally for 5 fitting epochs at the task boundary with a combination of embedding reconstruction, image-text alignment, teacher KL, and cross-entropy losses. When the next task $\mathcal{T}_{j+1}$ is introduced, client $k$ samples $G_k^{(j)}$ to synthesize a small buffer of replay tuples $\mathcal{B}_k^{(j)} = \{(\hat{\mathbf{h}}_{V,i}, \hat{\mathbf{h}}_{T,i}, \hat{y}_i)\}_{i=1}^{|\mathcal{B}|}$. Since $G_k^{(j)}$, the generated replay buffer, and the Fisher statistics remain local, PPFR follows an operational raw-data-free deployment model rather than claiming a formal differential-privacy guarantee. The generated buffer $\mathcal{B}_k^{(j)}$ is then interleaved with the new task's data $\mathcal{D}_k^{(j+1)}$ for local training. Including these synthetic embedding-level samples in the training process, as reflected in \cref{eq:overall_local_loss}, allows the model to rehearse prior event knowledge while avoiding storage or transmission of raw multimodal posts.
We formalize the replay synthesis process as
\begin{align}
\resizebox{0.85\columnwidth}{!}{$\displaystyle
\mathcal{B}_k^{(j)} = \left\{(\hat{\mathbf{h}}_{V,i}, \hat{\mathbf{h}}_{T,i}, \hat{y}_i)\ \middle|\ (\hat{\mathbf{h}}_{V,i}, \hat{\mathbf{h}}_{T,i}, \hat{y}_i) \sim G_k^{(j)}(\mathbf{z}_i, c_i),\ \mathbf{z}_i \sim \mathcal{N}(\mathbf{0}, \mathbf{I}) \right\}_{i=1}^{|\mathcal{B}|}
$},
\label{eq:replay_buffer}
\end{align}
where $c_i$ is the sampled local event code used to condition replay generation.
We define the replay-augmented dataset by
\begin{align}
\tilde{\mathcal{D}}_k^{(j+1)} = \mathcal{D}_k^{(j+1)} \cup \mathcal{B}_k^{(j)}.
\label{eq:augmented_dataset}
\end{align}
Semantic drift can still arise when replay tuples synthesized at task $\mathcal{T}_j$ are reused several task boundaries later, because the downstream adapters continue to change while the generator summarizes only past local evidence. Task-boundary refitting, teacher-smoothed replay labels, and the joint embedding-alignment losses reduce that mismatch by reanchoring the replay tuples to the local representation geometry seen at the time of generation. Residual drift is nevertheless possible on long task sequences, so PPFR should be interpreted as a compact rehearsal anchor rather than as a lossless memory of earlier multimodal posts.

\subsection{Task-Similarity-Aware Gradient Aggregation}
In federated continual learning, client drift is exacerbated because the statistical heterogeneity across clients is compounded by temporal heterogeneity. At any given time, different clients might be training on different tasks or on data that represents different facets of the same task, leading to local updates that may conflict with each other. A naive federated averaging of these updates can lead to destructive interference, harming the global model's performance on both old and new tasks.
To address this, we introduce TSGA, a server-side mechanism to modulate the aggregation process. The core idea is to identify and down-weight client updates that are outliers, as they are likely to be either off-topic for the current learning objective or of low quality. After receiving the parameter updates $\Delta \mathbf{w}_k = \mathbf{w}_k - \mathbf{w}_{\text{g}}$ from each participating client $k$ in a given round, the server first computes the average update direction:
\begin{align}
\bar{\mathbf{g}} = \frac{1}{|\mathcal{S}|} \sum_{k \in \mathcal{S}} \frac{\Delta \mathbf{w}_k}{\|\Delta \mathbf{w}_k\|},
\label{eq:avg_grad}
\end{align}
where $\mathcal{S}$ is the set of participating clients. The server then measures the alignment of each client's update with this average direction using cosine similarity:
\begin{align}
s_k = \frac{\Delta \mathbf{w}_k \cdot \bar{\mathbf{g}}}{\|\Delta \mathbf{w}_k\| \|\bar{\mathbf{g}}\|}.
\label{eq:cosine_sim}
\end{align}
Clients whose updates are poorly aligned with the consensus direction (i.e., have a low similarity score $s_k$) are likely to be sources of interference. We introduce a similarity threshold $\tau$. Any client with $s_k < \tau$ is considered an outlier and is excluded from the current round. The filtered participant set is
\begin{align}
\mathcal{S}' = \{k \in \mathcal{S} \mid s_k \ge \tau\},
\label{eq:filtered_clients}
\end{align}
and, if this set is empty, the round falls back to $\mathcal{S}'=\mathcal{S}$ to preserve a well-defined aggregation step even under highly conflicting two-client rounds. The filtered set is then used with normalized aggregation weights
\begin{align}
\alpha_k = \frac{N_k}{\sum_{i \in \mathcal{S}'} N_i}, \quad k \in \mathcal{S}',
\label{eq:aggregation_weight}
\end{align}
and the final global model is updated using a weighted average over only the clients in $\mathcal{S}'$:
\begin{align}
\mathbf{w}_{\text{g}} \leftarrow \mathbf{w}_{\text{g}} + \sum_{k \in \mathcal{S}'} \alpha_k \Delta \mathbf{w}_k.
\label{eq:tsga_aggregation}
\end{align}
By filtering out potentially disruptive updates, TSGA ensures a more stable and coherent learning trajectory for the global model, improving its ability to consolidate knowledge from diverse clients in a sequential learning environment.
The complete \texttt{FedCMM} workflow is summarized in \cref{alg:fedcmm_server} and \cref{alg:fedcmm_client}. The first algorithm captures the task-level server coordination, and the second details the client-side rehearsal and local update procedure. Round-level optimization involves only the clients sampled in a given communication round, whereas the task-boundary replay-generator fitting and Fisher estimation are local preparation steps executed by all clients after they receive the final global model for task $\mathcal{T}_j$.

\begin{algorithm}[!htbp]
\caption{\texttt{FedCMM} task-level server coordination.}
\label{alg:fedcmm_server}
\begin{minipage}{0.98\linewidth}
\begin{algorithmic}[1]
\State \textbf{Server executes:}
\State Initialize global model parameters $\mathbf{w}^{(0)}$
\For{each task $\mathcal{T}_j$ for $j=1, \dots, T$}
    \For{each round $r=1, \dots, R_j$}
        \State $\mathcal{S}_r \leftarrow$ Select a random subset of $C$ clients
        \State Broadcast $\mathbf{w}^{(j)}_{r-1}$ to all clients in $\mathcal{S}_r$
        \State Initialize empty list of updates $\mathcal{U} \leftarrow []$
        \For{each client $k \in \mathcal{S}_r$ \textbf{in parallel}}
            \State $\Delta \mathbf{w}_k \leftarrow \text{ClientUpdate}(k, \mathbf{w}^{(j)}_{r-1}, \mathcal{T}_j)$
            \State Add $\Delta \mathbf{w}_k$ to $\mathcal{U}$
        \EndFor
        \State // Aggregate with TSGA
        \State $\bar{\mathbf{g}} \leftarrow \frac{1}{|\mathcal{S}_r|} \sum_{k \in \mathcal{S}_r} \frac{\Delta \mathbf{w}_k}{\|\Delta \mathbf{w}_k\|}$
        \State $\mathcal{S}'_r \leftarrow \{k \in \mathcal{S}_r \mid \text{cosine}(\Delta \mathbf{w}_k, \bar{\mathbf{g}}) \ge \tau\}$
        \If{$\mathcal{S}'_r = \emptyset$}
            \State $\mathcal{S}'_r \leftarrow \mathcal{S}_r$
        \EndIf
        \State $\Delta \mathbf{w}^{(j)}_{r} \leftarrow \sum_{k \in \mathcal{S}'_r} \frac{N_k}{\sum_{i \in \mathcal{S}'_r} N_i} \Delta \mathbf{w}_k$
        \State $\mathbf{w}^{(j)}_{r} \leftarrow \mathbf{w}^{(j)}_{r-1} + \Delta \mathbf{w}^{(j)}_{r}$
    \EndFor
    \State $\mathbf{w}^{(j)} \leftarrow \mathbf{w}^{(j)}_{R_j}$
    \State \textbf{\\ Server signals all clients to prepare for the next task using the final global model:}
    \For{each client $k=1, \dots, K$ \textbf{in parallel}}
        \State Set client model to $\mathbf{w}^{(j)}$
        \State Train local generator $G_k^{(j)}$ on its data $\mathcal{D}_k^{(j)}$
        \State Compute FIMs $\mathbf{F}_{M,k}^{(j)}$ on data $\mathcal{D}_k^{(j)}$
        \State Store $G_k^{(j)}$ and FIMs locally for use in task $\mathcal{T}_{j+1}$
    \EndFor
\EndFor
\State \Return $\mathbf{w}^{(T)}$
\end{algorithmic}

\end{minipage}
\end{algorithm}

\begin{algorithm}[t]
\caption{\texttt{FedCMM} client rehearsal and local update.}
\label{alg:fedcmm_client}
\begin{algorithmic}[1]
\Procedure{ClientUpdate}{$k, \mathbf{w}, \mathcal{T}_j$}
\State Load local dataset $\mathcal{D}_k^{(j)}$ for task $\mathcal{T}_j$
\State // PPFR: Generate embedding-level replay tuples
\If{$j > 1$}
    \State // Load $G_k^{(j-1)}$ and $\mathbf{F}_{M,k}^{(j-1)}$ from previous task
    \State $\mathcal{B}_k^{(j-1)} \leftarrow$ Generate replay tuples using $G_k^{(j-1)}$
\Else
    \State $\mathcal{B}_k^{(j-1)} \leftarrow \emptyset$
\EndIf
\State // Local Training
\State Set local model to $\mathbf{w}_k \leftarrow \mathbf{w}$
\For{each local epoch $e=1, \dots, E$}
    \For{each batch in $\mathcal{D}_k^{(j)} \cup \mathcal{B}_k^{(j-1)}$}
        \State Compute loss $\mathcal{L}_k^{(j)}$ using \cref{eq:overall_local_loss} with $\mathbf{F}_{M,k}^{(j-1)}$
        \State Update $\mathbf{w}_k$ with gradient descent
    \EndFor
\EndFor
\State \Return $\Delta \mathbf{w}_k = \mathbf{w}_k - \mathbf{w}$
\EndProcedure
\end{algorithmic}
\end{algorithm}

\section{Experiments}
\label{sec:experiments}

Experiments were conducted on two federated continual social-event benchmarks in which every task is an image-text classification problem. The evaluation follows chronological event streams so that the model must preserve earlier event knowledge while adapting to later multimodal distributions.

\subsection{Datasets and Evaluation Metrics}
\noindent\textbf{Datasets and Metrics.} We use PHEME~\cite{kochkina2018allinone} and CrisisMMD~\cite{alam2018crisismmd}, two social-media event datasets that support image-text event understanding. PHEME contains rumor and breaking-news discussions collected from social media, with 6,273 text posts and 2,089 associated images across seven events. CrisisMMD contains crisis-related tweets from natural disasters, with 16,097 text instances and 18,126 images annotated for emergency-response categories. Following chronological event-incremental evaluation, events are ordered by time and grouped into three tasks with a $[2,2,3]$ class split for each dataset, yielding $T=3$ tasks per stream. All datasets are split into train/validation/test partitions with a 70/10/20 ratio. Unless otherwise noted, client data is partitioned across 10 clients with a Dirichlet label-skew distribution using concentration $\alpha=0.5$; we additionally report the severe non-IID setting $\alpha=0.1$. Let $A_{i,j}$ be the accuracy on task $\mathcal{T}_j$ after training up to task $\mathcal{T}_i$. We adopt three standard continual learning metrics: i) \textit{Average Accuracy (Acc)}, computed as $\frac{1}{T} \sum_{j=1}^T A_{T,j}$; ii) \textit{Backward Transfer (BWT)}, computed as $\frac{1}{T-1} \sum_{j=1}^{T-1} (A_{T,j} - A_{j,j})$ to quantify forgetting; and iii) \textit{Forward Transfer (FWT)}, computed as $\frac{1}{T-1} \sum_{j=2}^{T} A_{j-1,j}$. Higher Acc and FWT are better, and BWT is better when closer to zero.

\subsection{Baseline Methods and Implementation Details}
We compare \texttt{FedCMM} against a comprehensive set of baselines: \textit{Joint Training} (upper bound with all task data), \textit{Fine-tuning} (sequential lower bound), \textit{FedAvg-CL}~\cite{mcmahan2017communication}, \textit{FedProx-CL}~\cite{li2020proximal}, \textit{LwF-FL}~\cite{li2017learning}, \textit{EWC-FL}~\cite{kirkpatrick2017overcoming}, \textit{FedWeIT}~\cite{yoon2021fedweit}, and \textit{GEM-FL}~\cite{lopez2017gradient}. GEM-FL stores raw replay exemplars and is therefore included only as a privacy-violating rehearsal reference for comparison with direct exemplar replay, not as a deployable federated baseline under the same raw-data constraints. We also include AF-FCL~\cite{wuerkaixi2024accurate}, a generative-replay method based on accurate forgetting, and TEKNet-FL~\cite{qian2025learning}, a federated adaptation of the temporal event knowledge learner.
All experiments were implemented using PyTorch and Hugging Face Transformers. The base MLLM was \texttt{openbmb/MiniCPM-V-2.6-int4}. We employed PEFT with LoRA~\cite{hu2021lora} using rank $r=8$ and scaling factor 8. Each dataset is trained as a 3-task stream, with 50 federated rounds per task, 10 total clients, and 5 sampled clients per round unless otherwise stated. Local training uses paged AdamW with learning rate $2 \times 10^{-5}$, cosine decay, per-device batch size 1, and 16 gradient-accumulation steps. Each selected client performs one local epoch per federated round. PPFR fits only the local replay generator for 5 epochs at each task boundary, using a 64-dimensional Gaussian latent code and an embedding-level replay budget of $|\mathcal{B}|=100$. The replay weight is fixed to $\gamma=0.5$ on both PHEME and CrisisMMD, while the main stability hyperparameters are $\lambda=500$ for MA-EWC and $\tau=0.5$ for TSGA, selected to keep replay support, consolidation strength, and aggregation filtering on comparable scales. Experiments were conducted on NVIDIA L20 GPUs.

\begin{table*}[!t]
\caption{Performance comparison on 3-task PHEME and CrisisMMD under moderate and severe non-IID partitions. The values in parentheses indicate the Dirichlet heterogeneity parameter $\alpha$. Higher is better for Acc and FWT, while BWT is better when closer to zero. Best results are highlighted in \textbf{bold}. 
}
\label{tab:main_results}
\centering
\setlength{\tabcolsep}{2.7mm}
\resizebox{\textwidth}{!}{
\begin{tabular}{l|ccc|ccc|ccc|ccc}
\toprule
\rowcolor{gray!8}
\multirow{2}{*}{\textbf{Method}} & \multicolumn{3}{c|}{\textbf{PHEME (0.5)}} & \multicolumn{3}{c|}{\textbf{PHEME (0.1)}} & \multicolumn{3}{c|}{\textbf{CrisisMMD (0.5)}} & \multicolumn{3}{c}{\textbf{CrisisMMD (0.1)}} \\
\cmidrule(lr){2-4} \cmidrule(lr){5-7} \cmidrule(lr){8-10} \cmidrule(lr){11-13}
\rowcolor{gray!8}
 & Acc $\uparrow$ & BWT $\uparrow$ & FWT $\uparrow$ & Acc $\uparrow$ & BWT $\uparrow$ & FWT $\uparrow$ & Acc $\uparrow$ & BWT $\uparrow$ & FWT $\uparrow$ & Acc $\uparrow$ & BWT $\uparrow$ & FWT $\uparrow$ \\
\midrule
Joint Training & 89.2 & --- & --- & 89.2 & --- & --- & 91.4 & --- & --- & 91.4 & --- & --- \\
Fine-tuning & 68.4 & -29.1 & 0.4 & 65.7 & -32.4 & 0.0 & 62.6 & -30.8 & 0.5 & 56.1 & -36.8 & 0.4 \\
\midrule
FedAvg-CL~\cite{mcmahan2017communication} & 73.8 & -22.7 & 0.8 & 68.9 & -27.9 & 0.4 & 70.9 & -24.9 & 1.0 & 65.1 & -31.2 & 0.7 \\
FedProx-CL~\cite{li2020proximal} & 75.4 & -20.9 & 0.7 & 70.2 & -25.6 & 0.5 & 72.0 & -23.8 & 1.3 & 67.3 & -28.1 & 0.6 \\
LwF-FL~\cite{li2017learning} & 77.1 & -18.7 & 1.2 & 72.6 & -23.5 & 0.9 & 75.9 & -19.6 & 1.4 & 69.7 & -25.2 & 1.1 \\
EWC-FL~\cite{kirkpatrick2017overcoming} & 78.6 & -16.8 & 1.0 & 74.9 & -20.1 & 0.8 & 77.1 & -18.7 & 1.4 & 72.4 & -22.2 & 1.0 \\
FedWeIT~\cite{yoon2021fedweit} & 80.4 & -14.9 & 1.5 & 76.9 & -17.0 & 1.2 & 79.4 & -15.2 & 1.7 & 73.5 & -20.1 & 1.6 \\
GEM-FL~\cite{lopez2017gradient} & 83.2 & -12.0 & 1.9 & 79.6 & -14.5 & 1.2 & 82.2 & -12.0 & 2.2 & 76.9 & -16.0 & 1.9 \\
AF-FCL~\cite{wuerkaixi2024accurate} & 84.1 & -10.9 & 1.8 & 79.4 & -14.9 & 1.7 & 82.6 & -11.2 & 2.1 & 78.1 & -14.4 & 1.7 \\
TEKNet-FL~\cite{qian2025learning} & 84.6 & -10.2 & \textbf{2.3} & 80.7 & -13.0 & 1.6 & 83.4 & -10.5 & 2.3 & 78.0 & -14.7 & \textbf{2.4} \\
\midrule
\rowcolor{lightkeycolor}
\texttt{FedCMM} (Ours) & \textbf{86.3} & \textbf{-8.7} & 2.0 & \textbf{81.8} & \textbf{-12.6} & \textbf{1.9} & \textbf{84.4} & \textbf{-8.8} & \textbf{2.5} & \textbf{79.6} & \textbf{-12.8} & 2.1 \\
\bottomrule
\end{tabular}
}
\vspace{-8px}
\end{table*}

\subsection{Main Results}
\cref{tab:main_results} reports the comparison under moderate and severe heterogeneity. On PHEME with $\alpha=0.5$, \texttt{FedCMM} reaches 86.3\% Acc, improving over TEKNet-FL by 1.7\% and over AF-FCL by 2.2\%. The gain is not limited to the final event group: BWT improves from -10.2\% for TEKNet-FL to -8.7\%, indicating less damage to earlier events. Under the severe $\alpha=0.1$ split, all methods lose accuracy, but the ranking is not a simple shifted copy of the moderate case: GEM-FL has slightly higher Acc than AF-FCL, whereas AF-FCL has stronger FWT. \texttt{FedCMM} retains a 1.1\% Acc lead and keeps BWT 0.4\% closer to zero than TEKNet-FL. On CrisisMMD, \texttt{FedCMM} obtains 84.4\% and 79.6\% under $\alpha=0.5$ and $\alpha=0.1$, respectively, below the non-continual Joint Training upper bound but above the strongest continual baselines. Several close comparisons are dataset-dependent; for example, AF-FCL is slightly stronger than TEKNet-FL in severe CrisisMMD Acc, while TEKNet-FL gives better FWT on severe CrisisMMD and moderate PHEME. Accordingly, the most stable advantage of \texttt{FedCMM} lies in retention-oriented metrics, especially BWT, rather than in achieving the highest FWT on every dataset split. GEM-FL is competitive because raw exemplars provide direct rehearsal, yet it stores original samples and therefore serves only as a privacy-violating rehearsal reference rather than a deployable federated option. The gaps are therefore not uniform across datasets or metrics: CrisisMMD benefits more from visually grounded evidence, while PHEME depends more on event-specific textual shifts and stance cues. Across both streams, the strongest improvements appear on BWT, suggesting that the proposed modules prioritize forgetting reduction while preserving useful, but not uniformly maximal, forward plasticity for later crisis and rumor events.

\subsection{Ablation Studies}
To validate each component's contribution, we conducted ablation studies on PHEME under $\alpha=0.5$. \cref{tab:ablation_components} removes each component individually, and \cref{tab:ablation_maewc} analyzes the MA-EWC design.

\noindent\textbf{Contribution of each component.} As shown in \cref{tab:ablation_components}, FedAvg-CL reaches 73.8\% Acc with a BWT of -22.7\%, indicating severe forgetting under event-incremental PHEME. Adding all three components increases Acc to 86.3\% and improves BWT to -8.7\%. The component effects are not identical across metrics. Removing PPFR causes the largest retention loss, reducing Acc to 82.4\% and worsening BWT to -13.6\%. That 4.9-point BWT degradation is larger than the corresponding drops from removing MA-EWC or TSGA, which directly supports the interpretation that replay is the primary anchor for preserving old-event decision boundaries once the stream advances to later events. Removing MA-EWC gives 83.7\% Acc and weaker FWT, suggesting that replay alone cannot fully stabilize modality-specific adapters across task switches. Removing TSGA gives 84.5\% Acc; its BWT is slightly better than removing MA-EWC, but the full method still improves both retention and transfer. Notably, the nonparallel pattern across metrics is expected because replay, consolidation, and aggregation filtering affect different failure modes. PPFR dominates when old-event boundaries are most vulnerable, MA-EWC becomes more important when parameter drift accumulates in the vision, language, and projector adapters, and TSGA becomes more important when cross-client updates conflict at the server. The ablation pattern therefore shows that no single component fully explains the final performance.

\begin{table}[t]
\caption{Result of component ablation on PHEME ($\alpha=0.5$).}
\label{tab:ablation_components}
\centering
\setlength{\tabcolsep}{3.5pt}
\resizebox{\columnwidth}{!}{
\begin{tabular}{l|ccc|ccc}
\toprule
\rowcolor{gray!8}
\textbf{Configuration} & \textbf{MA-EWC} & \textbf{PPFR} & \textbf{TSGA} & \textbf{Acc $\uparrow$} & \textbf{BWT $\uparrow$} & \textbf{FWT $\uparrow$} \\
\midrule
FedAvg-CL~\cite{mcmahan2017communication} & \xmark & \xmark & \xmark & 73.8 & -22.7 & 0.8 \\
\texttt{FedCMM}$_{\text{w/o-MA}}$ & \xmark & \cmark & \cmark & 83.7 & -10.9 & 1.7 \\
\texttt{FedCMM}$_{\text{w/o-R}}$ & \cmark & \xmark & \cmark & 82.4 & -13.6 & 1.6 \\
\texttt{FedCMM}$_{\text{w/o-G}}$ & \cmark & \cmark & \xmark & 84.5 & -9.5 & 1.9 \\
\midrule
\rowcolor{lightkeycolor}
\texttt{FedCMM} & \cmark & \cmark & \cmark & \textbf{86.3} & \textbf{-8.7} & \textbf{2.0} \\
\bottomrule
\end{tabular}
}
\vspace{-15px}
\end{table}

\noindent\textbf{Analysis of MA-EWC.} \cref{tab:ablation_maewc} shows that the Fisher penalty must respect the MLLM module structure. Removing EWC entirely gives 82.8\% Acc and -12.4\% BWT. A uniform EWC penalty improves this to 84.3\%, but still trails the full modality-aware design. Single-branch protection is unstable and metric-dependent: vision-only protection reaches 83.1\% Acc but gives relatively high FWT, while language-only protection gives higher Acc and BWT but lower FWT. Notably, visual stabilization appears to help early adaptation to new events, whereas language-side protection better preserves accumulated rumor cues. The full model separately protects the vision encoder, language backbone, and projector, reaching 86.3\% and the best BWT of -8.7\%. In turn, the evidence supports the claim that forgetting in social-event MLLM adaptation is distributed across visual, textual, and cross-modal alignment modules. The separate Fisher design is therefore not only a stronger regularizer, but also a better match to how multimodal adapters store task-specific evidence.

\begin{table}[t]
\caption{Result of MA-EWC ablation on PHEME ($\alpha=0.5$).}
\label{tab:ablation_maewc}
\centering
\setlength{\tabcolsep}{2.8mm}
\resizebox{\columnwidth}{!}{
\begin{tabular}{l|cccc|ccc}
\toprule
\rowcolor{gray!8}
\textbf{Configuration} & \textbf{V} & \textbf{L} & \textbf{P} & \textbf{Sep.} & \textbf{Acc $\uparrow$} & \textbf{BWT $\uparrow$} & \textbf{FWT $\uparrow$} \\
\midrule
\texttt{FedCMM}$_{\text{no-EWC}}$ & \xmark & \xmark & \xmark & \xmark & 82.8 & -12.4 & 1.5 \\
\texttt{FedCMM}$_{\text{uni-EWC}}$ & \cmark & \cmark & \cmark & \xmark & 84.3 & -10.9 & 1.7 \\
\texttt{FedCMM}$_{\text{V}}$ & \cmark & \xmark & \xmark & \xmark & 83.1 & -11.7 & 1.8 \\
\texttt{FedCMM}$_{\text{L}}$ & \xmark & \cmark & \xmark & \xmark & 83.9 & -10.8 & 1.5 \\
\midrule
\rowcolor{lightkeycolor}
\texttt{FedCMM} & \cmark & \cmark & \cmark & \cmark & \textbf{86.3} & \textbf{-8.7} & \textbf{2.0} \\
\bottomrule
\end{tabular}
}
\fontsize{7.2pt}{8pt}\selectfont {
 \begin{minipage}{\columnwidth}
   \vspace{1px}
   \quad \textit{Notes}: V = Vision Encoder, L = Language Backbone, P = Projector, Sep. = Separate Fisher Matrices.
 \end{minipage}
}
\vspace{-15px}
\end{table}

\begin{figure*}[!t]
    \centering
    \includegraphics[width=\textwidth]{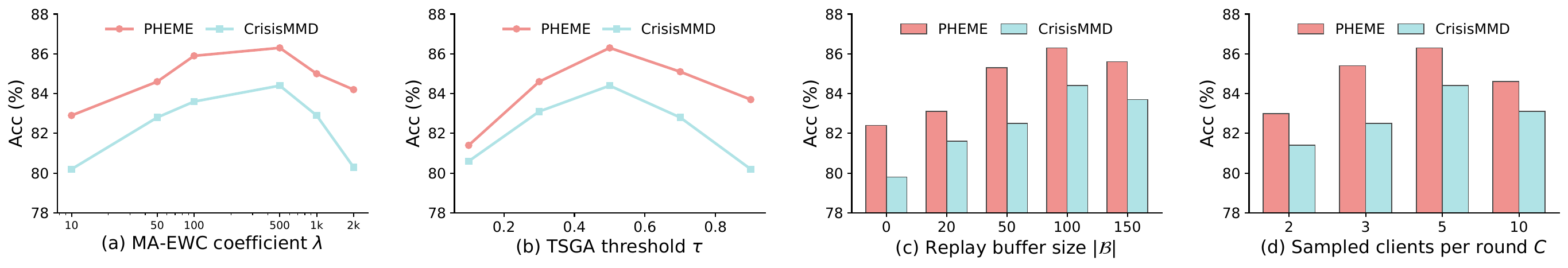}
    \caption{Parameter sensitivity of \texttt{FedCMM} on PHEME and CrisisMMD, where the four panels analyze hyperparameters $\lambda$, $\tau$, $|\mathcal{B}|$, and $C$.}
    \label{fig:analysis_plots}
\vspace{-8px}
\end{figure*}

\subsection{Parameter Sensitivity and Robustness}
\noindent\textbf{Sensitivity to hyperparameters.} \cref{fig:analysis_plots} shows sensitivity analyses for four key hyperparameters. The curves are not perfectly symmetric because the two benchmarks fail in different ways: PHEME is dominated by cross-event semantic shift, while CrisisMMD is more affected by noisy image-text alignment and class imbalance. For MA-EWC strength $\lambda$, both datasets peak at $\lambda=500$, but the decline after the peak is sharper on CrisisMMD. Increasing $\lambda$ from 500 to 2000 reduces CrisisMMD from 84.4\% to 80.3\%, whereas PHEME decreases from 86.3\% to 84.2\%. In particular, over-constraining adapters can suppress crisis-event adaptation even when it protects old tasks. For the TSGA threshold $\tau$, 0.5 gives the best result, but the neighboring points are uneven: $\tau=0.3$ is more competitive on CrisisMMD (83.1\%), whereas PHEME is less sensitive at $\tau=0.7$. The replay buffer trend is also nonuniform. Moving from no replay to $|\mathcal{B}|=100$ improves PHEME by 3.9\% and CrisisMMD by 4.6\%, but the intermediate gains differ across datasets, and increasing to 150 does not help. Consequently, low-quality synthetic tuples can begin to dilute current-task supervision. Finally, sampling $C=5$ clients per round is best, while the $C=3$ and $C=10$ gaps differ across datasets. In aggregate, the pattern across all four parameters indicates that the default configuration is not an isolated lucky point; it lies in a stable region where consolidation, replay, and aggregation filtering remain balanced.

\begin{figure}[!t]
    \centering
    \includegraphics[width=\columnwidth]{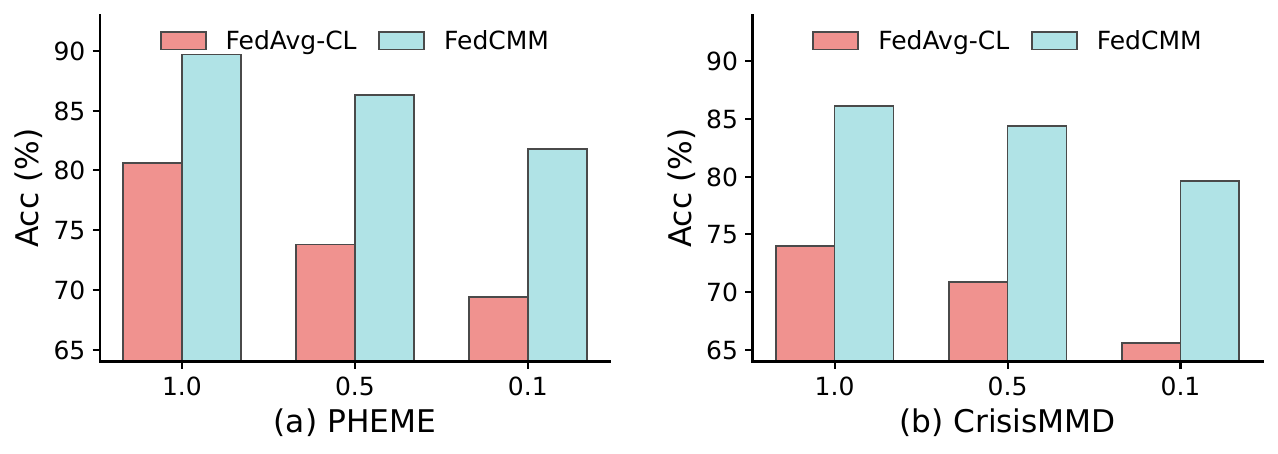}
    \caption{Robustness to client heterogeneity on PHEME and CrisisMMD.}
    \label{fig:heterogeneity_plots}
\vspace{-8px}
\end{figure}

\noindent\textbf{Robustness to data heterogeneity.} \cref{fig:heterogeneity_plots} confirms that \texttt{FedCMM} remains more stable as the client label distribution becomes skewed. On PHEME, FedAvg-CL decreases from 80.6\% at $\alpha=1.0$ to 69.4\% at $\alpha=0.1$, while \texttt{FedCMM} decreases from 89.7\% to 81.8\%. Notably, the gap is not constant: it is 9.1\% under the mild split, 12.5\% at $\alpha=0.5$, and 12.4\% under severe heterogeneity. On CrisisMMD, FedAvg-CL falls from 74.0\% to 65.6\%, whereas \texttt{FedCMM} falls from 86.1\% to 79.6\%. A related asymmetry appears in the degradation pattern: the moderate-to-severe drop is larger than the mild-to-moderate drop, especially for FedAvg-CL on CrisisMMD. Disaster events appear to have more heterogeneous visual evidence across clients. Importantly, the proposed method does not eliminate non-IID sensitivity; rather, it reduces the additional forgetting caused by skewed client updates. In turn, the results support the role of TSGA as a stabilizer rather than as a substitute for better client coverage. The remaining drop at $\alpha=0.1$ also shows that the evaluation is not saturated, leaving visible room for stronger personalization or adaptive client sampling.

\noindent\textbf{Robustness to task ordering.} \cref{tab:robustness_task_order} evaluates desynchronized client progression on PHEME, where 30\% of clients lag one task behind at each boundary. Notably, the setting is harsher than ordinary label skew because stale clients optimize an earlier event distribution while synchronized clients have already moved to the next group. FedAvg-CL drops from 73.8\% to 65.7\%, and EWC-FL drops from 78.6\% to 71.4\%. \texttt{FedCMM} drops from 86.3\% to 82.4\%, losing less than 4\%. A smaller degradation is consistent with the interaction between TSGA and PPFR: stale updates tend to have lower agreement with the round consensus, and replay keeps the retained clients from over-specializing to only the latest event. Beyond a simple heterogeneity test, the asynchronous setting specifically probes whether aggregation can tolerate clients being at different points in the continual stream. Moreover, the property is important for social-event deployments, where clients may receive delayed event evidence or skip event groups entirely.

\begin{table}[t]
\caption{Accuracy of task-order robustness on PHEME. $\Delta$ denotes the accuracy drop from synchronized to asynchronous progression.}
\label{tab:robustness_task_order}
\centering
\setlength{\tabcolsep}{6mm}
\resizebox{\columnwidth}{!}{
\begin{tabular}{l|ccc}
\toprule
\rowcolor{gray!8}
\textbf{Method} & \textbf{Sync. $\uparrow$} & \textbf{Async. $\uparrow$} & $\boldsymbol{\Delta \uparrow}$ \\
\midrule
FedAvg-CL~\cite{mcmahan2017communication} & 73.8 & 65.7 & -8.1 \\
EWC-FL~\cite{kirkpatrick2017overcoming} & 78.6 & 71.4 & -7.2 \\
\midrule
\rowcolor{lightkeycolor}
\texttt{FedCMM} & \textbf{86.3} & \textbf{82.4} & \textbf{-3.9} \\
\bottomrule
\end{tabular}
}
\vspace{-8px}
\end{table}

\subsection{Local-Epoch Sensitivity and Efficiency Analysis}
\begin{figure}[!t]
    \centering
    \includegraphics[width=\columnwidth]{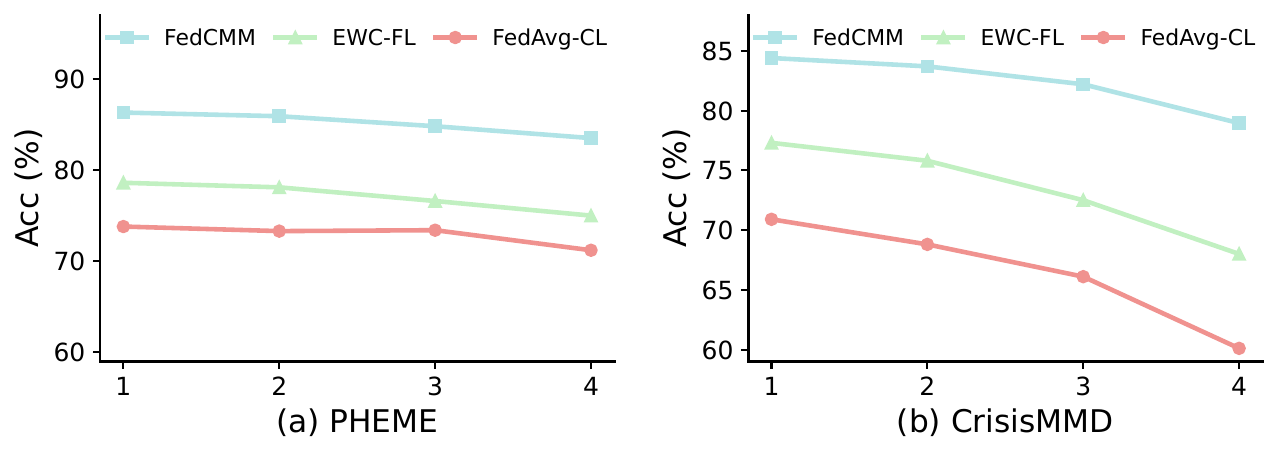}
    \caption{Effect of local epoch number $E$. Larger $E$ amplifies both cross-client non-IID drift and temporal drift across evolving event groups.}
    \label{fig:local_epoch_plots}
\vspace{-8px}
\end{figure}

\noindent\textbf{Effect of local epoch number $E$.} \cref{fig:local_epoch_plots} shows average accuracy as $E$ increases from 1 to 4. All methods peak at $E=1$, confirming that aggressive local training amplifies client drift in temporally skewed social-event streams. On PHEME, FedAvg-CL drops from 73.8\% to 71.2\%, EWC-FL drops from 78.6\% to 75.0\%, and \texttt{FedCMM} drops from 86.3\% to 83.5\%. On CrisisMMD, the absolute decline is larger for the baselines but not by a fixed margin: FedAvg-CL loses 10.8\%, EWC-FL loses 9.3\%, and \texttt{FedCMM} loses 5.45\%. The intermediate points also differ; for example, FedAvg-CL on PHEME is relatively flat from $E=2$ to $E=3$ before dropping at $E=4$, whereas all methods on CrisisMMD show a steeper and more consistent decline. Longer local training is especially harmful because it deepens client adaptation to each local non-IID partition while the task stream itself is evolving, so ordinary federated client drift is compounded by temporal drift across event groups. 
In particular, the nonparallel degradation shows that \texttt{FedCMM} is not merely an upward-shifted baseline. Replay and aggregation filtering reduce the damage caused by longer client-side optimization, but they do not make large local epochs harmless. For this reason, the default setting uses one local epoch for the MLLM update, while PPFR's five fitting epochs apply only to the small replay generator at task boundaries. The result also suggests that increasing local computation is not a reliable substitute for communication in federated continual MLLM adaptation.

\noindent\textbf{Convergence efficiency.} \cref{tab:efficiency} reports the number of federated rounds needed for the aggregated global model to first reach 50\% and 75\% test accuracy after a communication round, together with the final global-model accuracy at round 50, under $\alpha=0.5$. The milestones are therefore measured on round-level global checkpoints rather than by averaging client-local states. At the 50\% milestone, \texttt{FedCMM} converges in 8 rounds on PHEME and 10 rounds on CrisisMMD, achieving speedups of approximately 2.25$\times$ over FedAvg-CL (18 and 23 rounds) and 1.6$\times$ over EWC-FL (13 and 15 rounds). At the more demanding 75\% target, \texttt{FedCMM} requires 13 and 19 rounds, while EWC-FL needs 28 and 33 rounds, roughly twice as many; FedAvg-CL cannot reach 75\% within the 50-round budget on either benchmark, saturating below that threshold. At the final checkpoint (round 50), \texttt{FedCMM} achieves 86.3\% on PHEME and 84.4\% on CrisisMMD, exceeding EWC-FL by 7.7\% and 7.3\%, respectively.

\begin{table}[t]
\caption{Federated rounds for the global model to first reach 50\% ($R_{50}\downarrow$) and 75\% ($R_{75}\downarrow$) accuracy after a communication round, and final global-model accuracy at round 50 (Final$\uparrow$) on PHEME and CrisisMMD.}
\label{tab:efficiency}
\centering
\setlength{\tabcolsep}{2mm}
\resizebox{\columnwidth}{!}{
\begin{tabular}{l|ccc|ccc}
\toprule
\rowcolor{gray!8}
\multirow{2}{*}{\textbf{Method}} & \multicolumn{3}{c|}{\textbf{PHEME}} & \multicolumn{3}{c}{\textbf{CrisisMMD}} \\
\cmidrule(lr){2-4} \cmidrule(lr){5-7}
\rowcolor{gray!8}
 & $R_{50}\downarrow$ & $R_{75}\downarrow$ & Final $\uparrow$ & $R_{50}\downarrow$ & $R_{75}\downarrow$ & Final $\uparrow$ \\
\midrule
Joint Training & --- & --- & 89.2 & --- & --- & 91.4 \\
\midrule
FedAvg-CL~\cite{mcmahan2017communication} & 18 & --- & 73.8 & 23 & --- & 70.9 \\
EWC-FL~\cite{kirkpatrick2017overcoming} & 13 & 28 & 78.6 & 15 & 33 & 77.1 \\
\midrule
\rowcolor{lightkeycolor}
\texttt{FedCMM} & \textbf{8} & \textbf{13} & \textbf{86.3} & \textbf{10} & \textbf{19} & \textbf{84.4} \\
\bottomrule
\end{tabular}
}
\vspace{-8px}
\end{table}

\subsection{Discussion}
Superior performance of \texttt{FedCMM} across both benchmarks stems from the coordinated integration of three complementary mechanisms rather than from any single component in isolation: i) \textit{Modality-aware regularization} via MA-EWC applies module-specific Fisher penalties to the vision encoder, language backbone, and cross-modal projector independently, preventing the uniform parameter shrinkage that causes a single-branch EWC to under-protect whichever modality updates most aggressively in the current task; ii) \textit{Privacy-preserving replay} via PPFR restores sample-space supervision without exposing raw client data, and the empirical gap between PPFR and standard exemplar replay (GEM-FL) confirms that synthetic embeddings preserve enough decision geometry to achieve comparable retention; and iii) \textit{Similarity-aware aggregation} via TSGA filters out client updates that are directionally misaligned with the consensus, reducing the amplification of task-ordering drift that ordinary FedAvg cannot suppress. In this sense, the main novelty lies in multimodal federated continual integration, where rehearsal, consolidation, and aggregation control are coupled to address the same evolving task stream. Nevertheless, several limitations warrant attention: optimal $\lambda$ and $\tau$ values are currently determined by grid search rather than adaptive scheduling; the Fisher diagonal approximation may underestimate parameter importance in cross-modal projector layers where interference is most nonlinear; replay embeddings can accumulate semantic drift across longer task sequences even when local generator refitting is used; and scaling to $T\!\geq\!6$ tasks has not been validated. Future work will pursue learnable hyperparameter schedules~\cite{wang2024comprehensive}, more expressive Fisher surrogates for low-rank adapters, stronger replay-refresh strategies, broader people-centric activity recognition streams~\cite{liu2023distributional}, and extended evaluation on the seven-task PHEME$\rightarrow$CrisisMMD combined stream~\cite{qian2025learning,dong2024federated}.

\section{Conclusion}
\label{sec:conclusion}

In this paper, we address catastrophic forgetting in federated fine-tuning of Multimodal Large Language Models on evolving social-event streams. The proposed \texttt{FedCMM} framework combines modality-aware parameter regularization, raw-data-free embedding-level replay, and similarity-aware gradient aggregation to preserve prior knowledge during sequential adaptation. Extensive experiments show that \texttt{FedCMM} consistently achieves the strongest average accuracy and BWT, while remaining competitive on FWT. Its robustness to data heterogeneity and asynchronous task ordering indicates that the framework is suitable for privacy-sensitive distributed multimodal monitoring settings such as content moderation and crisis understanding. Future work will focus on improving efficiency and adaptive hyperparameter scheduling for longer continual federated task.

\bibliographystyle{IEEEtran}
\begin{spacing}{0.95}
  \bibliography{refs.bib}
\end{spacing}
\end{document}